\documentclass[11pt]{article}

\usepackage{acl}

\usepackage{times}
\usepackage{latexsym}

\usepackage{booktabs}

\usepackage{amsmath}

\usepackage[T1]{fontenc}

\usepackage[utf8]{inputenc}

\usepackage{subcaption}

\usepackage{microtype}

\usepackage{inconsolata}

\usepackage{graphicx}

\title{Bilingual Language Models Capture Cross-Linguistic Influence in Second-Language Reading}

\author{
  \textbf{Abderrahmane Issam}$^{1}$,
  \textbf{Irene E. Winther}$^{2}$,
  \textbf{Stefan L. Frank}$^{2}$ \\
  $^{1}$Maastricht University,
  $^{2}$Radboud University \\
  \texttt{abderrahmane.issam@maastrichtuniversity.nl} \\
  \texttt{\{irene.winther, stefan.frank\}@ru.nl}
}

\begin{document}
\maketitle
\begin{abstract}
While monolingual language model (LM) surprisal is widely established as a predictor of reading times, it cannot capture the non-native (L2) reading dynamics that are shaped by the reader's first language (L1). In this work, we investigate whether the cross-lingual transfer within bilingual models aligns with human L1-driven biases during L2 reading. We evaluate bilingual models on eye-tracking data of readers from 12 diverse L1 backgrounds reading English texts. To isolate L1-specific transfer, we compare bilingual models trained on an L1 that matches the readers' L1 background to both a monolingual English baseline and bilingual models trained on other L1s. We demonstrate that matching the readers' L1 improves the prediction of L2 reading times. This predictive advantage is concentrated in late-stage eye-tracking measures, modulated by L1-L2 script overlap, and peaks among intermediate-proficiency readers. These findings provide evidence that bilingual LMs capture aspects of cross-linguistic influence, and that such influence is present in L2 reading behavior.
\end{abstract}

\section{Introduction}

Language Models (LMs) are increasingly being investigated for their alignment with human cognitive processing \cite{ARMENI2017579, karamolegkou-etal-2023-mapping, niu2026, Futrell_Mahowald_2026}. Within this domain, a prominent line of work evaluates LMs on eye-tracking data \cite{DEMBERG2008193, stefan2009, goodkind-bicknell-2018-predictive, oh-schuler-2023-surprisal, shain2024, wilcox-etal-2023-testing}, resting on the well-established assumption that fixation durations reflect cognitive effort during sentence processing \cite{rayner1998, schotter2025}. Under surprisal theory \cite{hale-2001-probabilistic, LEVY20081126}, the cognitive effort of processing a word is tied its predictability in context (surprisal). While substantial work shows that LM surprisal successfully predicts native-language (L1) reading times \cite{SMITH2013302, goodkind-bicknell-2018-predictive, wilcox2020predictivepowerneurallanguage, shain2024}, much less is known regarding the ability of bilingual LMs to predict reading behavior in a non-native language (L2).

One candidate source of difference between native and non-native reading is cross-linguistic influence from the reader's L1 \cite{Kim_2025, aleman-banon2021, Odlin_1989, Tokowicz_MacWhinney_2005, tokowicz2024competition, foucart_language_2021}. If such influence actively shapes L2 reading, a monolingual LM trained solely on the target language cannot, by definition, capture it.

In parallel, research in NLP demonstrates that multilingual models learn shared cross-lingual representations \cite{pires-etal-2019-multilingual, libovicky-etal-2020-language, varda-marelli-2023-data}, enabling zero-shot performance transfer across languages \cite{wu-dredze-2019-beto, conneau-etal-2020-emerging, conneau-etal-2020-unsupervised}. However, it remains an open question whether these representations translate into predictive biases that align with human L2 reading behavior. 

In this work, we investigate whether bilingual LMs' cross-lingual transfer aligns with human cross-linguistic influence during L2 reading. We evaluate bilingual GPT-2 models on the Multilingual Eye Movement Corpus (MECO L2) \cite{Kuperman2025}, using eye-tracking recordings from participants across 12 diverse L1 backgrounds reading English texts. To examine whether the duration of L1 training affects L2 prediction in the models, we train bilingual models across multiple Step of Exposure (SoE) stages, varying the moment during L1 learning at which L2 English is introduced. Hence, we manipulate the duration of initial L1 training and the amount of L2 exposure. We evaluate model surprisal across four reading-time metrics (first fixation duration, gaze duration, go-past time, and total fixation duration) to distinguish early lexical access from late-stage structural reanalysis and text integration \cite{rayner1998, liversedge2000, INHOFF1984612}. 

Our investigation addresses two primary questions. First, we establish whether bilingual models' surprisal explains variance in L2 reading times that a monolingual baseline cannot. Second, we compare L1-matched bilingual models (where the model's initial training language matches the reader's L1) to L1-mismatched controls (where the model is trained on a different L1) to determine whether this predictive advantage is driven by language-specific transfer rather than a generic multilingual effect.

We find a bilingual predictive advantage concentrated in late-stage reading measures, with little to no advantage in early ones. Furthermore, L1-matched models outperform mismatched models, indicating that the explained variance in L2 reading times is L1-specific. Finally, we observe that this advantage is modulated by L1-L2 script overlap and peaks for L2 readers with intermediate proficiency. In summary, these findings provide evidence that bilingual LMs are capable of modeling cross-linguistic influence, and that such influence is present in human L2 reading behavior.

\section{Related Works}
\subsection{Bilingual Models for L2 Acquisition}
The success of multilingual language models has prompted research into whether their internal cross-lingual transfer mechanisms align with human cross-linguistic influence \cite{mueller-etal-2020-cross, papadimitriou-jurafsky-2020-learning, oba-etal-2023-second, aoyama-schneider-2024-modeling, yadavalli-etal-2023-slabert, issam2026studycrosslinguisticinfluencelanguage, salhan2026beetlebilingualmodelsuite}. To simulate the dynamics of L2 acquisition, previous studies have modeled bilingual exposure by training on unbalanced L1 and L2 data \cite{dhar-bisazza-2021-understanding, winther_2021}, introducing the L2 sequentially \cite{winther_2021, constantinescu-etal-2025-investigating, arnett-etal-2025-acquisition, issam2026studycrosslinguisticinfluencelanguage, salhan2026beetlebilingualmodelsuite}, freezing L1 representations during L2 exposure \cite{papadimitriou-jurafsky-2020-learning, yadavalli-etal-2023-slabert, aoyama-schneider-2024-modeling}, or using specialized training objectives \cite{oba-etal-2023-second}. Previous research on bilingual LMs has investigated various effects of bilingual exposure, including syntactic transfer \cite{oba-etal-2023-second, yadavalli-etal-2023-slabert, issam2026studycrosslinguisticinfluencelanguage}, cross-linguistic priming \cite{frank2021cross, michaelov-etal-2023-structural, arnett-etal-2025-acquisition}, and L1 influence on  L2 reading times \cite{aoyama-schneider-2024-modeling, salhan2026beetlebilingualmodelsuite}. Building on this, we adopt the sequential training strategy of \citet{issam2026studycrosslinguisticinfluencelanguage} to train bilingual models for several L1s, and  evaluate their ability to predict human L2 reading times.
\vspace{-0.15cm}
\subsection{Evaluating LMs on L2 Reading Times}
While most work evaluating language models on eye-tracking data focuses on L1 reading \cite{DEMBERG2008193, stefan2009, goodkind-bicknell-2018-predictive, oh-schuler-2023-surprisal, shain2024, wilcox-etal-2023-testing}, the release of bilingual speaker corpora such as CELER \cite{berzak2022} and MECO L2 \cite{Kuperman2025} has enabled the study of non-native processing \cite{de-varda-marelli-2022-effects, berzak2023, aoyama-schneider-2024-modeling, salhan2026beetlebilingualmodelsuite}. Most of this work, however, still relies on surprisal from monolingual models for the prediction of L2 reading times.

A smaller body of work has explored bilingual models for this purpose, but critical gaps remain. For instance, in their evaluation of reading times, \citet{salhan2026beetlebilingualmodelsuite} lacked monolingual baselines and L1-mismatched controls, making it impossible to isolate L1-specific transfer. While \citet{aoyama-schneider-2024-modeling} included a monolingual baseline, they relied on frozen L1 representations and compared independent model fits rather than testing for additional variance explained. Furthermore, both studies omitted L1 control groups and restricted their evaluations primarily to early reading metrics like gaze duration. Our approach systematically addresses these limitations. First, we train fully unfrozen models, allowing bilingual representations to dynamically interact. Second, we measure each bilingual model's  variance explained over a monolingual baseline and test whether L1-matched surprisal explains variance beyond L1-mismatched surprisal within the same regression analysis. Empirically, we broaden the scope by evaluating L1 and L2 readers across four eye-tracking measures and 12 L1s, further examining how this predictive advantage is modulated by reader proficiency and script overlap.

\section{Methods}
\subsection{Bilingual LM Training}
We follow the training approach of \citet{issam2026studycrosslinguisticinfluencelanguage} to train bilingual GPT-2 LMs for each L1-L2 language pair. We acquire multilingual data from FineWeb \cite{Penedo2024}, using the 10-billion token sample (\texttt{sample-10BT}) for English and the language-specific splits from FineWeb-v2 \cite{penedo2025fineweb2pipelinescale} for the L1s. We follow the approach of \citet{issam2026studycrosslinguisticinfluencelanguage} to minimize code-switching by filtering out sentences using sentence- and word-level language detection. While completely mitigating code-switching remains an open challenge in NLP \cite{burchell-etal-2024-code}, this filtering step is crucial. Because code-switching is known to enhance cross-lingual transfer \cite{ijcai2020p533, wang-etal-2025-investigating-scaling}, leaving it unfiltered could inflate the observed L1 influence and contaminate the monolingual baseline.

We vary the Step of Exposure (SoE) as the step at which we introduce the L2 to study how extended L1 training and delayed L2 exposure affect the prediction of L2 reading data. Increasing the SoE results in longer initial L1 training and less L2 training. Since we train all models for the same total amount of training steps, this mimics a comparison between bilingual speakers of the same age but with different ages of first exposure to the L2. Once the L2 is introduced, we continue training on the L1 data by interleaving L1 and L2 sequences. This prevents catastrophic forgetting and maintains the L1 influence on the L2 \cite{constantinescu-etal-2025-investigating, arnett-etal-2025-acquisition}. Furthermore, it mirrors a realistic scenario where non-native speakers maintain strong knowledge of their L1, either due to critical period effects \cite{Lenneberg01121967, bylund2019} or persistent L1 use during L2 learning.

We train a SentencePiece tokenizer \cite{kudo-richardson-2018-sentencepiece} for each L1-L2 pair using a balanced subset of 10 million lines (5 million lines per language). Once trained, this tokenizer is used to process the full corpus of each language independently, packing the text into shuffled sequences of 256 tokens. Finally, from these fully tokenized corpora, we sample one billion tokens per language for LM training and one million tokens for evaluation.

We train all models for a total of 64k training steps. During monolingual training, batches consists of monolingual sequences, while bilingual training consists of interleaved L1 and L2 sequences in each batch. We train 4 SoE models for each L1-L2 pair by initiating bilingual training at training step 0, 16k, 32k and 48k. We use a fixed learning rate with 5K warmup similarly to \citet{issam2026studycrosslinguisticinfluencelanguage} to avoid confounding our variables, because a linear learning rate would lead to later SoE models being trained on L2 with a lower learning rate. Finally, besides the bilingual models, we train a monolingual baseline using an identical tokenizer to each bilingual pair exclusively on English data. In total, 55 GPT-2 models were trained in this study (1 monolingual and 4 bilingual SoE models for each of the 11 L1-L2 pairs). Detailed training hyperparameters are provided in Appendix \ref{sec:hyperparameters}.


\subsection{Evaluation Framework}
We use MECO L2 \cite{Kuperman2025}, which provides reading data of English as a second language. The dataset includes eye-tracking data of participants from 13 L1 language backgrounds collected across 16 institutions. Rather than aggregating participants by L1, we keep each institution as a distinct subset. We include all L1 groups except the Basque subset, which we exclude because of insufficient data representation in FineWeb-v2. We also include the native English (En-UK) subset to serve as a native control group. Participants across all subsets read the same 12 English texts sourced from standardized reading assessments. See Appendix \ref{sec:languages} for further details on the subsets.

We evaluate model predictions against four eye-tracking measures:
\begin{enumerate}
\item \textit{First Fixation Duration:} The duration of the first fixation on a word.
\item \textit{Gaze Duration:} The sum of all fixation durations landing on a word before the eye leaves it for the first time.
\item \textit{Go-Past Time:} The total time elapsed from the initial fixation on a word until the reader's gaze moves to the right of that word. This includes all time spent regressing to earlier parts of the text.
\item \textit{Total Fixation Duration:} The sum of all fixation durations on a word, including subsequent fixations if the eye returns to it.
\end{enumerate}
These measures are thought to reflect progressively later stages of cognitive processing \cite{rayner1998, liversedge2000}. Early measures, such as first fixation duration and gaze duration, are assumed to capture initial lexical access, including word form recognition and meaning retrieval. Late measures, such as go-past time and total fixation duration, are thought to reflect post-lexical processing, such as structural reanalysis and syntactic or semantic integration. Separating these metrics allows us to distinguish context-invariant word recognition from higher-level, context-dependent processing \cite{INHOFF1984612, berzak2023}.

We derive surprisal estimates from every LM trained for this study: 1 monolingual and 4 SoE bilingual LMs for each of the 11 L1-L2 language pairs (excluding native English and Basque subsets). We calculate the surprisal of each word as its negative log-probability. For words segmented into multiple subwords by the tokenizer, we sum the surprisals of the constituent tokens.

To evaluate the predictions, we fit linear mixed-effects models (LMMs) with crossed random intercepts for participants and words. The baseline fixed effects include log-frequency, word length, and word position in the sentence. We also include the log-frequency and length of the preceding word ($w_{i-1}$) to account for spillover effects. Word frequencies are extracted using the \texttt{wordfreq} library \cite{robyn_speer_2022_7199437}.

\subsubsection{Model Comparisons}
We test whether bilingual models capture cross-linguistic cognitive processing through two main comparisons.

\paragraph{Bilingual vs. Monolingual}
First, we compare each bilingual model against the monolingual English baseline. Our hypothesis is that if bilingual models capture cross-linguistic influence, their surprisal estimates will yield better predictions of non-native reading times than a strictly monolingual model's. To account for inherent differences in baseline predictability across the different language subsets, we use a pseudo-$R^2$ metric:
\begin{equation}
R_{\text{pseudo}}^2 = 1 - \frac{LL_{\text{model}}}{LL_{\text{baseline}}}
\end{equation}
where $LL_\text{baseline}$ is the log-likelihood of an LMM containing only the surface baseline features. We then calculate $\Delta R^2=R_\text{bilingual}^2-R_\text{monolingual}^2$ to measure the proportion of explainable variance gained by the bilingual model. We evaluate the LMs on the respective MECO L1 subsets and the native En-UK. The native subset serves as a control group to verify that any observed advantage of a bilingual model is specific to L2 reading rather than a general improved fit to reading times.

\paragraph{Matched vs. Mismatched Bilingual Models}

Second, we compare matched bilingual LMs to mismatched bilingual LMs. A bilingual LM is considered matched if its L1 matches the native language of the evaluated MECO subset; all other bilingual LMs are considered mismatched. If bilingual LMs develop L1-specific transfer effects similar to those of human readers, matched LMs should predict reading times more accurately than mismatched LMs. To quantify this difference, we calculate the difference in log-likelihood ($\Delta LL$) between two non-nested LMMs: one using surprisal from a matched bilingual LM and the other using surprisal from a mismatched LM. 

To compute formal $p$-values, we also conduct nested regression model comparisons. We fit an encompassing model ($m_{\text{both}}$) containing both the matched and mismatched surprisal predictors. We then compare this encompassing model against restricted models containing either the matched or the mismatched predictor alone. All three models use identical baseline covariates and random effects structures, differing solely in their surprisal terms. The unique explanatory variance added by matched surprisal is calculated as $LL(m_{\text{both}}) - LL(m_{\text{mismatch}})$. Because $m_\text{mismatch}$ is nested within $m_\text{both}$, we evaluate statistical significance of this added variance using a likelihood ratio test. 

\subsection{Proficiency Modulation Analysis}
\label{sec:methods_proficiency}

To evaluate how reader proficiency modulates the predictive advantage of bilingual surprisal over monolingual baselines, we employ a two-stage regression inspired by \citet{berzak2023}. In the first stage, we isolate reader-specific surprisal sensitivity across eye-tracking measures by fitting individual participant regressions that control for baseline features. For each reader, we compute the bilingual explanatory advantage ($\Delta R^2 = R^2_{\text{bilingual}} - R^2_{\text{monolingual}}$). In the second stage, we model $\Delta R^2$ as a function of reader L2 English proficiency, measured via LexTALE \cite{lemhofer2011}. To ensure orthogonal linear and quadratic predictors, proficiency scores are mean-centered. We estimate parameters using Weighted Least Squares (WLS), where each reader is weighted by their total observation count. 

\begin{figure*}[t]
    \centering
    \includegraphics[width=\textwidth]{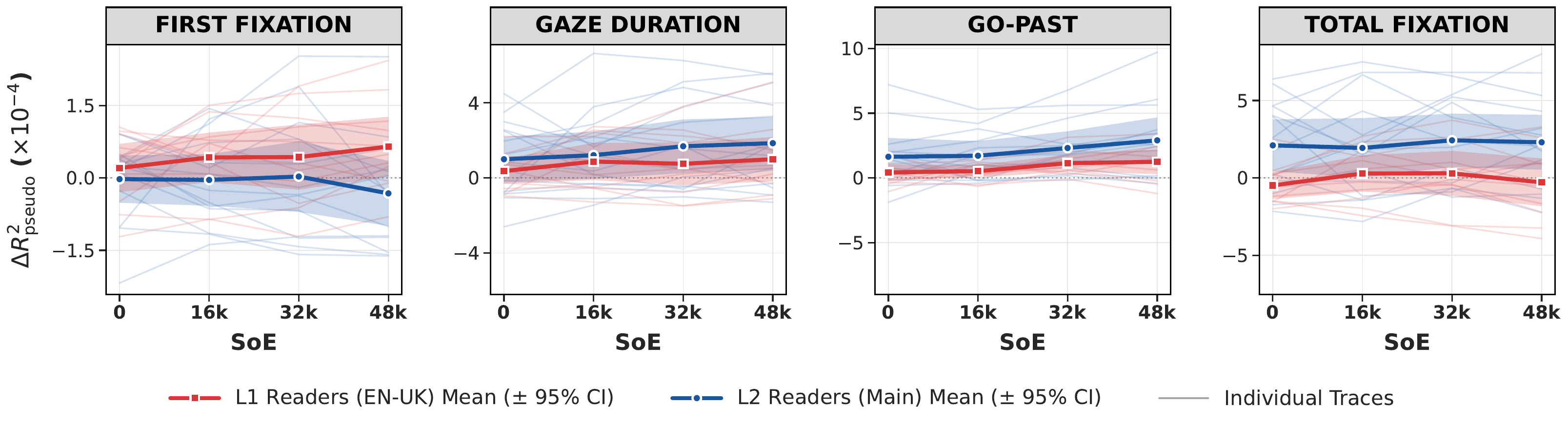}
    \caption{Difference in explained variance $\Delta R_{\text{pseudo}}^2$ (scaled by $10^4$) in four reading measures as a function of SoE. The blue line represents the mean across all L2 reader subsets, while the red line denotes the native English (En-UK) control group. Faint lines indicate individual model evaluations. Positive values indicate a predictive advantage for the bilingual surprisal. With the exception of first fixation duration, bilingual surprisal consistently provides greater predictive power for L2 than L1 readers.}
    \label{fig:bi_vs_mono}
\end{figure*}

To test whether proficiency moderation is linear or non-linear, we compare a linear model ($\Delta R^2 \sim \text{prof}$) with a quadratic model ($\Delta R^2 \sim \text{prof} + \text{prof}^2$). We select between them using $\Delta AIC$ and test the quadratic coefficient using HC3 heteroskedasticity-consistent standard errors \cite{MACKINNON1985305}. To rule out confounding by differences across L1 subsets, we then refit the quadratic model with L1-subset fixed effects and standard errors clustered by L1 subset, and check whether the quadratic term remains significant.

\section{Results and Analysis}
\subsection{Are L2 Models Better at Predicting L2 Data than Monolingual Models?}
\label{sec:bi_mono}

Figure \ref{fig:bi_vs_mono} shows the difference in explained variance ($\Delta R_{pseudo}^2$) between the matched bilingual models and the monolingual baselines across four SoE stages. The results show a difference between early and late reading measures. 

For first fixation duration, the bilingual surprisal provides no predictive advantage for L2 readers. The mean $\Delta R_{pseudo}^2$ for L2 readers starts near zero and drops slightly below zero at the SoE=48k. In contrast, for the later reading measures, the bilingual model yields a clear advantage. In both gaze duration and go-past time, the L2 mean is strictly positive and increases consistently with SoE. In total fixation time, the L2 mean remains positive and stable across all SoEs. 

Crucially, across gaze duration, go-past time, and total fixation duration, the native English control group stays close to zero and consistently below the L2 mean. This gap confirms that the bilingual models capture variance specific to non-native reading behavior, rather than offering a generic predictive improvement for all readers. 

\subsection{Are Matched Models Better Than Mismatched Models?}
\label{sec:match_mismatched}
\begin{figure*}[t]
    \centering
    \includegraphics[width=\textwidth]{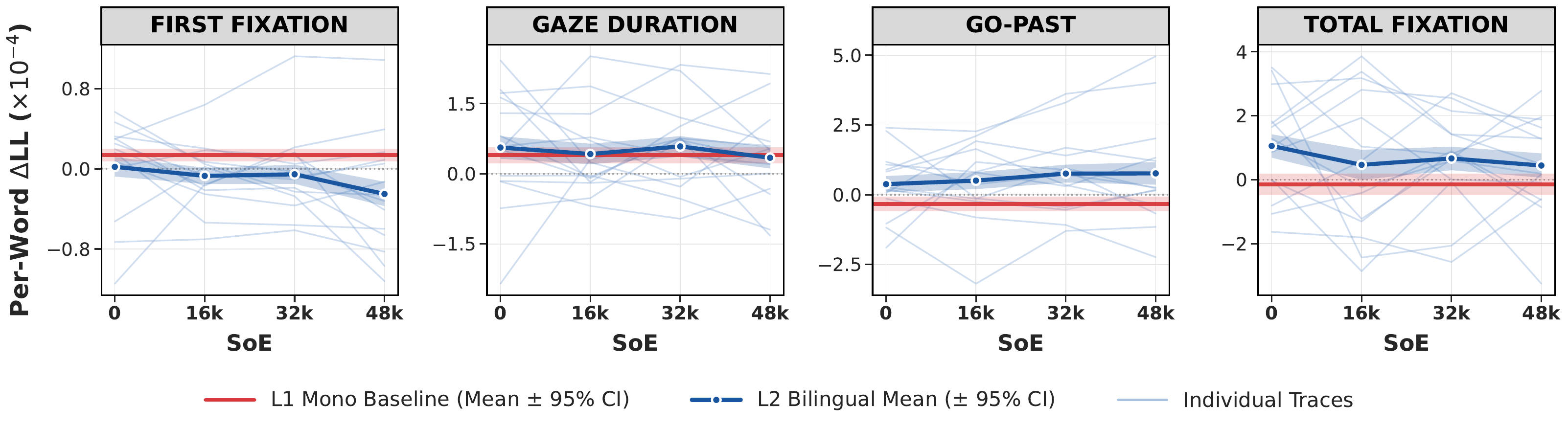}
    \caption{Per-word $\Delta$LL (scaled by $10^{4}$) between matched and mismatched models in four reading measures as a function of SoE. Solid lines and shaded bands represent subset means with 95\% confidence intervals for the bilingual models (blue) and monolingual baseline (red), with faint blue lines indicating individual subset evaluations. Higher values indicate a stronger predictive advantage for the matched model relative the mismatched one. With the exception of first fixation duration, bilingual models match or outperform the monolingual baseline.}
    \label{fig:dll_metrics_all}
\end{figure*}
Figure \ref{fig:dll_metrics_all} illustrates the non-nested per-word $\Delta LL$ between LMMs using matched and mismatched bilingual surprisals. To control for evaluation biases or tokenization artifacts, we include a monolingual baseline derived by treating each L1 monolingual LM as matched and comparing it against the other monolingual LMs. If the matched advantage is genuinely driven by bilingual cross-linguistic transfer, it should exceed this strictly monolingual baseline. Except for first fixation duration, the matched bilingual models consistently outperform their mismatched counterparts, and this advantage strengthens in later processing measures. Notably, go-past time is the only metric exhibiting a clear upward trend in matched model advantage as SoE increases. Although there is variance among individual subset traces (representing the mean per-word $\Delta LL$ between matched and mismatched models for each L1 subset), the matched bilingual models generally exceed the monolingual control.

The nested model results in Table \ref{tab:nested_dll} show whether this matched advantage provides unique explanatory variance. By evaluating both predictors simultaneously, we can assess which model contributes a greater share of unique information based on their mean $\Delta LL$ and significance ratios. These nested comparisons perfectly align with our non-nested findings. During early lexical access (first fixation duration), matched models offer no L1-specific predictive advantage, contributing less unique variance than mismatched models. In contrast, across all later reading measures (gaze duration, go-past time, and total fixation duration), the matched models contribute substantially more unique log-likelihood and achieve statistical significance at much higher rates than their mismatched counterparts.

Ultimately, even though the surprisal estimates from different bilingual models are highly correlated (see Appendix \ref{sec:surprisal_corr} for surprisal correlations), the matched models consistently explain significantly more unique variance during late-stage reading. These results indicate that bilingual LMs successfully capture aspects of L1-specific structural processing.

\begin{table}[t]
\centering
\scriptsize
\begin{tabular}{lcccc}
\toprule
\textbf{Metric} & \multicolumn{2}{c}{\textbf{Mean $\Delta\text{LL}$}} & \multicolumn{2}{c}{\textbf{Sig. Ratio ($p < 0.05$)}} \\
\cmidrule(lr){2-3} \cmidrule(lr){4-5}
& \textbf{Match} & \textbf{Mism.} & \textbf{Match} & \textbf{Mism.} \\
\midrule
First Fixation  & 1.07 & \textbf{1.33} & 17.0\% & \textbf{23.2\%} \\
Gaze Duration   & \textbf{3.23} & 1.78 & \textbf{55.0\%} & 30.4\% \\
Go-Past    & \textbf{3.32} & 1.69 & \textbf{53.7\%} & 30.8\% \\
Total Fixation  & \textbf{6.53} & 4.00 & \textbf{77.3\%} & 64.3\% \\
\bottomrule
\end{tabular}
\caption{Nested model comparisons ($m_{\text{both}}$) evaluating the unique contribution of matched versus mismatched bilingual surprisal averaged across all SoE models and language subsets. With the exception of First Fixation, matched models' mean $\Delta LL$ and significance ratio are higher than the mismatched models.}
\label{tab:nested_dll}
\end{table}

\subsection{The Role of Script Overlap}
\label{sec:script}
A primary structural dimension separating the evaluated subsets is writing system overlap with English. We examine this effect by partitioning the results from Figure \ref{fig:dll_metrics_all} by script. The Latin script group consists of 8 MECO subsets across 7 L1s, while the Non-Latin group consists of 6 subsets across 4 L1s (with the exception of go-past time, where the \texttt{ch\_s} subset is missing). See Appendix \ref{sec:languages} for details on the L1 subsets and their scripts.

This analysis is motivated by prior NLP research showing that shared scripts facilitate cross-lingual transfer in LMs \cite{dhamecha-etal-2021-role, moosa-etal-2023-transliteration}, as well as psycholinguistic studies examining the processing dynamics of reading across different writing systems \cite{miwa2014reading, Taylor_Mukai_2023, arab2026}.

\begin{figure*}[t]
    \centering
    \includegraphics[width=\textwidth]{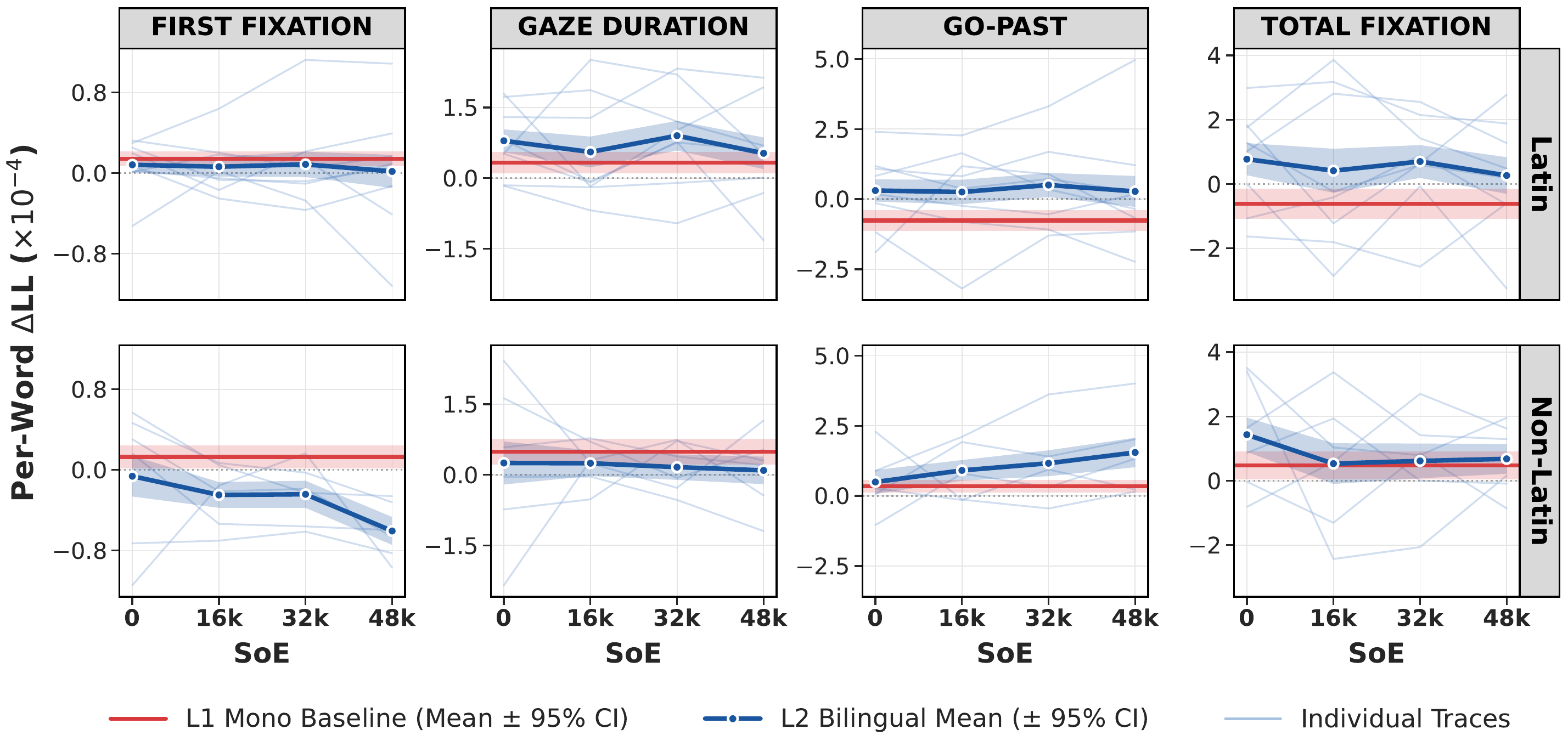}
    \caption{Per-word $\Delta$LL (scaled by $10^{4}$) between matched and mismatched models in four reading measures as a function of SoE, grouped by Latin (top) and Non-Latin (bottom) L1 scripts. Solid lines and shaded bands denote subset means with 95\% confidence intervals for the bilingual models (blue) and monolingual baseline (red). Faint blue lines indicate individual subset traces. Early measures (First Fixation and Gaze Duration) reveal a clear divergence in predictive behavior between the two script groups.}
    \label{fig:dll_script}
\end{figure*}

Figure \ref{fig:dll_script} illustrates the resulting per-word $\Delta LL$. In first fixation duration, neither script group displays a bilingual advantage. Latin bilingual models hover near zero close to the monolingual baseline, while Non-Latin bilingual models decline sharply below zero as SoE increases, dropping well below their within-group baseline. In gaze duration, the groups diverge. Latin bilingual models remain consistently positive and comfortably above the monolingual baseline, whereas Non-Latin models hover near zero and below the baseline across all SoEs. For later measures (go-past time and total fixation duration), however, bilingual models in both groups perform comparably to or better than the monolingual baseline. Non-Latin subsets also display a pronounced, steady upward trend in go-past time as SoE increases.


\subsection{Effect of Reader Proficiency}
\label{sec:proficiency_analysis}
\begin{figure*}[t]
  \centering
  \includegraphics[width=\textwidth]{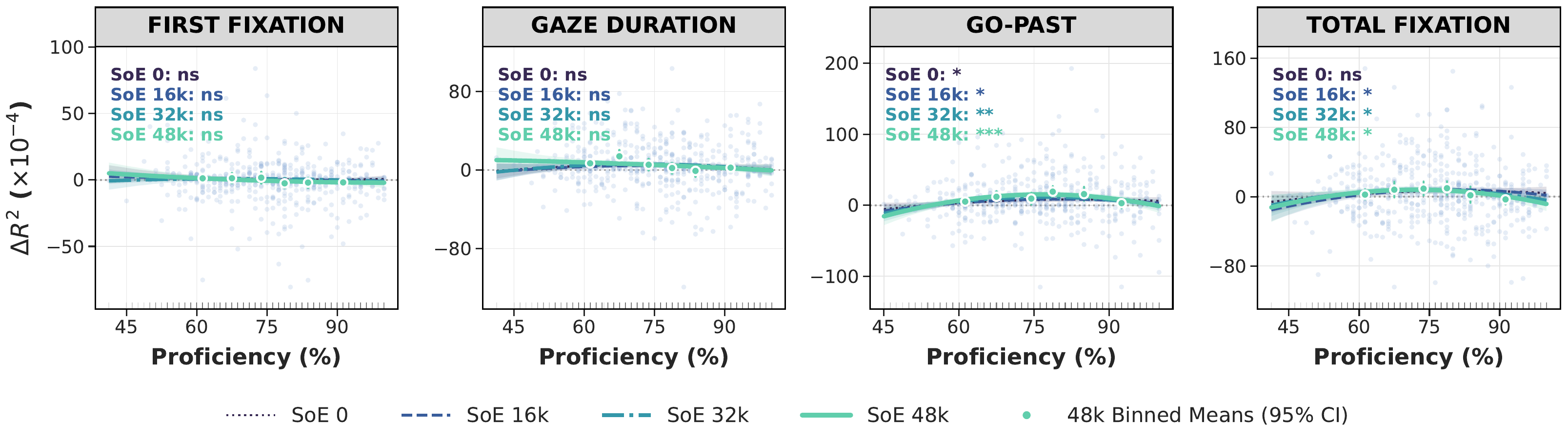}
  \caption{Bilingual surprisal advantage $\Delta R^2$ (scaled by $10^4$) across readers' English proficiency (LexTALE \%) in four reading measures. Curves show quadratic fits per SoE model; circular markers denote binned means ($\pm 95\%$ CI) at SoE=48k. Faint dots indicate individual observations, and the horizontal dotted line denotes zero gain. Inset labels indicate the significance of the quadratic proficiency term in a WLS with L1-subset fixed effects and standard errors clustered by L1 subset ($^*p < 0.05$, $^{**}p < 0.01$, $^{***}p < 0.001$, $^{\text{ns}}\text{not significant}$). While early measures show negligible non-linearity, late measures (Go-Past, Total Fixation) show an inverted-U trajectory peaking among intermediate-proficiency readers ($65\text{--}80\%$).}
  \label{fig:l2_r2diff}
\end{figure*}

Previous work demonstrates that L2 proficiency non-monotonically modulates the predictive power of monolingual surprisal \cite{berzak2023}. Inspired by this finding, we ask a novel question: how does the added predictive power of bilingual surprisal change with reader proficiency? We hypothesize an inverted-U relationship for this bilingual predictive advantage over the monolingual baseline. This advantage should be minimal at both extremes of the proficiency spectrum: beginners rely primarily on shallow, context-independent features such as word frequency \cite{Diependaele2013, berzak2023}, while advanced readers no longer depend on L1 transfer as their reading behavior approaches that of a native speaker \cite{Chen_Fang_Perfetti_2026}. Consequently, the bilingual advantage should peak for intermediate readers, who possess enough fluency to engage in contextual prediction but still rely heavily on L1 syntactic biases during processing.

As illustrated in Figure~\ref{fig:l2_r2diff}, while early lexical access measures maintain a linear relationship (all $p_{\text{quad}} > 0.05$), late processing measures (go-past time and total fixation duration) display a pronounced inverted-U trajectory that strongly favors the quadratic polynomial ($\Delta\text{AIC} > 2.0$ for SoEs $\ge$ 16k; Appendix~\ref{sec:quadratic_estimation}). Crucially, this curvature remains robust: go-past time shows significant quadratic modulation across all SoE ($0.001 < p_{\text{quad}} < 0.05$), and total fixation duration reaches significance for SoE $\ge$ 16k ($p_{\text{quad}} < 0.05$). The bilingual advantage peaks for intermediate readers ($65\text{--}80\%$ LexTALE) and descends toward zero for the most advanced readers ($> 90\%$ LexTALE).

Our hypothesis of an inverted-U trajectory is thus confirmed, though it applies exclusively to late measures where contextual integration and syntactic reanalysis are prominent. Importantly, this pattern is not an artifact of LexTALE: evaluating a composite proficiency index extracted via Principal Component Analysis (PCA) across all five proficiency tests in MECO (see Appendix~\ref{sec:proficiency_pca}) replicates this trajectory, confirming robust significance of non-monotonic moderation in go-past time (all SoEs) and total fixation duration (SoE=16k).

To contextualize this result, Appendix~\ref{sec:absolute_surprisal} visualizes the absolute explanatory power ($R^2$) of each model independently. As expected, both models exhibit an inverted-U trend across all metrics, corroborating \citet{berzak2023} by showing that general contextual prediction peaks for intermediate learners. Crucially, however, Figure~\ref{fig:l2_r2diff} demonstrates that the differential advantage of the bilingual model ($\Delta R^2$) follows this non-linear trajectory strictly during late measures. This indicates that while intermediate readers generally rely on broad contextual expectations, their late-stage processing is more strongly influenced by native-language biases.

\section{Discussion and Conclusion}
\label{sec:discussion}

In this work, we investigated whether bilingual LMs align with human cross-linguistic influence during L2 reading. Our findings demonstrate that bilingual LMs successfully capture aspects of L1-specific effects on L2 processing. Crucially, our two distinct evaluation frameworks---comparing bilingual LMs against a monolingual baseline (\S \ref{sec:bi_mono}), and contrasting L1-matched against L1-mismatched bilingual LMs (\S \ref{sec:match_mismatched})---yield consistent results. 

Across both comparisons, we observed a progressive shift in predictive performance from early to late reading measures. The predictive advantage of bilingual surprisals is absent in first fixation duration, begins to emerge in gaze duration, and becomes most pronounced in late-stage measures such as go-past time and total fixation duration. Regarding the SoE, we generally did not observe a uniform effect on predictability across all measures. This lack of a generalized trend may occur because the impact of extended L1 training varies considerably across different L1-L2 language pairs. The notable exception is go-past time, which exhibits a consistent upward trend in predictability as SoE increases.

Our analysis of writing system overlap (\S \ref{sec:script}) reveals a clear divergence in early processing. In gaze duration, only Latin L1 models showed a predictive advantage, whereas both script groups competed with or outperformed the monolingual baseline in late measures. These results can be interpreted through the Bilingual Interactive Activation (BIA+) model \cite{dijkstra2002}. When languages share a script, visual input can directly activate orthographic representations in both languages simultaneously, allowing cross-linguistic effects to emerge in early reading measures \cite{miwa2014reading}. In contrast, when scripts differ, direct orthographic co-activation is blocked, delaying cross-linguistic influence until downstream phonological or semantic integration occurs \cite{Taylor_Mukai_2023, miwa2014reading}. Consequently, cross-linguistic effects in Non-Latin groups emerge only in late-stage measures (go-past time and total fixation duration). While our findings cannot definitively confirm these specific cognitive mechanisms, this script-dependent processing divergence presents a compelling avenue for future investigation.

Additionally, our results reveal that the bilingual predictive advantage is non-linearly modulated by reader proficiency (\S \ref{sec:proficiency_analysis}). We observed an inverted-U trajectory that peaks for intermediate L2 readers and diminishes for highly advanced learners. Importantly, this proficiency modulation is robustly constrained to late processing measures. This indicates that while intermediate learners rely heavily on broad contextual expectations, their reliance on L1-driven biases surfaces predominantly during late-stage processing.

Across all findings, go-past time exhibits the most consistent alignment with L1-specific processing. Unlike the other reading measures that group eye fixations solely by location, go-past time groups fixations by sequence in time, directly measuring the time required to detect and resolve processing breakdowns \cite{LIVERSEDGE199855, liversedge2000, rayner1998}. By incorporating L1 structural patterns, the bilingual models successfully predict the regressive eye movements associated with L1-L2 conflicts, cautioning future research against relying solely on early reading measures.

In conclusion, our work bridges multilingual NLP and psycholinguistics by demonstrating that bilingual LMs internalize L1-specific predictive biases that mirror human non-native reading. Rather than a uniform advantage, we show that this cross-linguistic alignment is concentrated in late-stage structural processing and peaks among intermediate learners. Consequently, accurately modeling L2 reading requires evaluation frameworks that capture the full temporal complexity of non-native processing.


\section*{Limitations}
While we have attempted to rigorously evaluate bilingual LMs on L2 sentence processing, several limitations remain.

First, our evaluation is restricted to English as the target L2 and relies on a single eye-tracking corpus (MECO). Although MECO provides a rich diversity of L1 backgrounds, focusing exclusively on L2 English leaves it an open question whether our findings generalize to target languages with fundamentally different typological features, such as highly agglutinative languages or those with freer word order. Additionally, data collection for certain L1 subsets in MECO took place in regions where specific regional dialects or varieties of the language are spoken (e.g., Brazilian Portuguese). Because our bilingual models were trained on standard, general-domain corpora for each L1, this mismatch between standard pretraining text and regional variants may introduce noise and could explain some of the variation observed in individual model performance across subsets.

Second, while our script analysis groups languages based on their script overlap with English, this is only one axis of variation. Other typological factors, such as word order alignment and morphological similarity, likely affect L2 sentence processing. Future work should investigate how these broader structural overlaps independently modulate the predictive biases observed in L2 reading.

Third, our operationalization of reader proficiency relies on LexTALE, a metric shown to correlate with general L2 proficiency \cite{lemhofer2011} and widely adopted for this purpose (e.g., \citealt{Diependaele2013, cop2015, Caissie-Gilbert03072026}). However, its validity as a measure of global proficiency has been questioned \cite{Mayenco2023}, as it is fundamentally a lexical measure. To address this, we demonstrated the robustness of our results using a PCA-derived composite score based on all proficiency metrics reported in the MECO corpus. Nevertheless, these combined metrics still primarily index lower-level skills such as vocabulary size and decoding efficiency. As a result, the extent to which other proficiency factors such as higher-level comprehension or syntactic awareness modulate L1-driven predictive biases remains an open question.

Finally, we trained each model configuration using a single initialization seed. Although this precludes measuring variance across different weight initializations, we mitigate this limitation through the breadth of our experimental setup. Specifically, we trained four SoE bilingual models for each of the 11 L1s and compared them against corresponding monolingual baselines. By analyzing the consistency of these results across diverse languages and tracking how they evolve across SoEs, we ensure that the observed bilingual advantage is a robust trend rather than an artifact of a specific model checkpoint.

\bibliography{custom}

\appendix

\section{MECO Subsets and Languages}
\label{sec:languages}
Table \ref{tab:meco_samples} summarizes the L1 languages, writing scripts, total observations, and number of readers for each MECO L1 subset. While the exact sample sizes vary across individual reading measures due to missing eye-tracking data, every evaluated subset maintains a minimum of 7,639 samples per metric. The sole exception is the \texttt{ch\_s} subset, which completely lacks data for go-past time.

\begin{table}[ht]
\centering
\small
\setlength{\tabcolsep}{4pt}
\begin{tabular}{lllc}
\toprule
\textbf{Subset} & \textbf{Language \& Site} & \textbf{Samples}  & \textbf{Readers}\\
\midrule
\texttt{hi\_iiith} & Hindi (Hyderabad)* & 77,522 & 53 \\
\texttt{en\_uk}    & English (UK) & 66,041 & 50 \\
\texttt{bp}        & Brazilian Portuguese & 62,608 & 54 \\
\texttt{ge\_zu}    & German (Zurich) & 58,949 & 47 \\
\texttt{hi\_iitk}  & Hindi (Kanpur)* & 58,642 & 45 \\
\texttt{ru\_mo}    & Russian (Moscow)* & 57,817 & 49 \\
\texttt{ic}        & Icelandic & 56,821 & 45 \\
\texttt{ge\_po}    & German (Potsdam) & 49,604 & 43 \\
\texttt{sp\_ch}    & Spanish (Chile) & 49,434 & 45 \\
\texttt{ch\_s}     & Chinese simplified* & 47,685 & 46 \\
\texttt{se}        & Serbian* & 46,220 & 43 \\
\texttt{ch\_t}     & Chinese traditional* & 44,900 & 43 \\
\texttt{da}        & Danish & 25,815 & 25 \\
\texttt{no}        & Norwegian & 19,025 & 22 \\
\texttt{tr}        & Turkish & 13,797 & 14 \\
\bottomrule
\end{tabular}
\caption{MECO L2 subsets, corresponding L1 languages and recording site (* denotes Non-Latin), total number of samples (before filtering) and number of readers.}
\label{tab:meco_samples}
\end{table}

\section{Hyperparameters}
\label{sec:hyperparameters}

All models share the exact same GPT-2 architecture and training hyperparameters, which are detailed in Table \ref{tab:hyperparams}.

\begin{table}[ht]
\centering
\small
\begin{tabular}{lc}
\toprule
\textbf{Hyperparameter} & \textbf{Value} \\
\midrule
Layers & 12 \\
Embedding size & 768 \\
Hidden size & 768 \\
Intermediate hidden size & 3072 \\
Attention heads & 12 \\
Attention head size & 64 \\
Activation function & GELU \\
Vocab size & 50004 \\
Max sequence length & 256 \\
Position embedding & Absolute \\
Batch size & 64 \\
Train steps & 64k \\
Learning rate decay & Constant \\
Warmup steps & 5000 \\
Learning rate & $1 \times 10^{-4}$ \\
Adam $\epsilon$ & $1 \times 10^{-6}$ \\
Adam $\beta_1$ & 0.9 \\
Adam $\beta_2$ & 0.999 \\
Dropout & 0.1 \\
Attention dropout & 0.1 \\
Seed & 123 \\
\bottomrule
\end{tabular}
\caption{Architectural and optimization hyperparameters for the GPT-2 bilingual models.}
\label{tab:hyperparams}
\end{table}

\section{Bilingual Model Surprisal Correlation}
\label{sec:surprisal_corr}
\begin{figure}[h]
    \centering
    \includegraphics[width=0.49\textwidth]{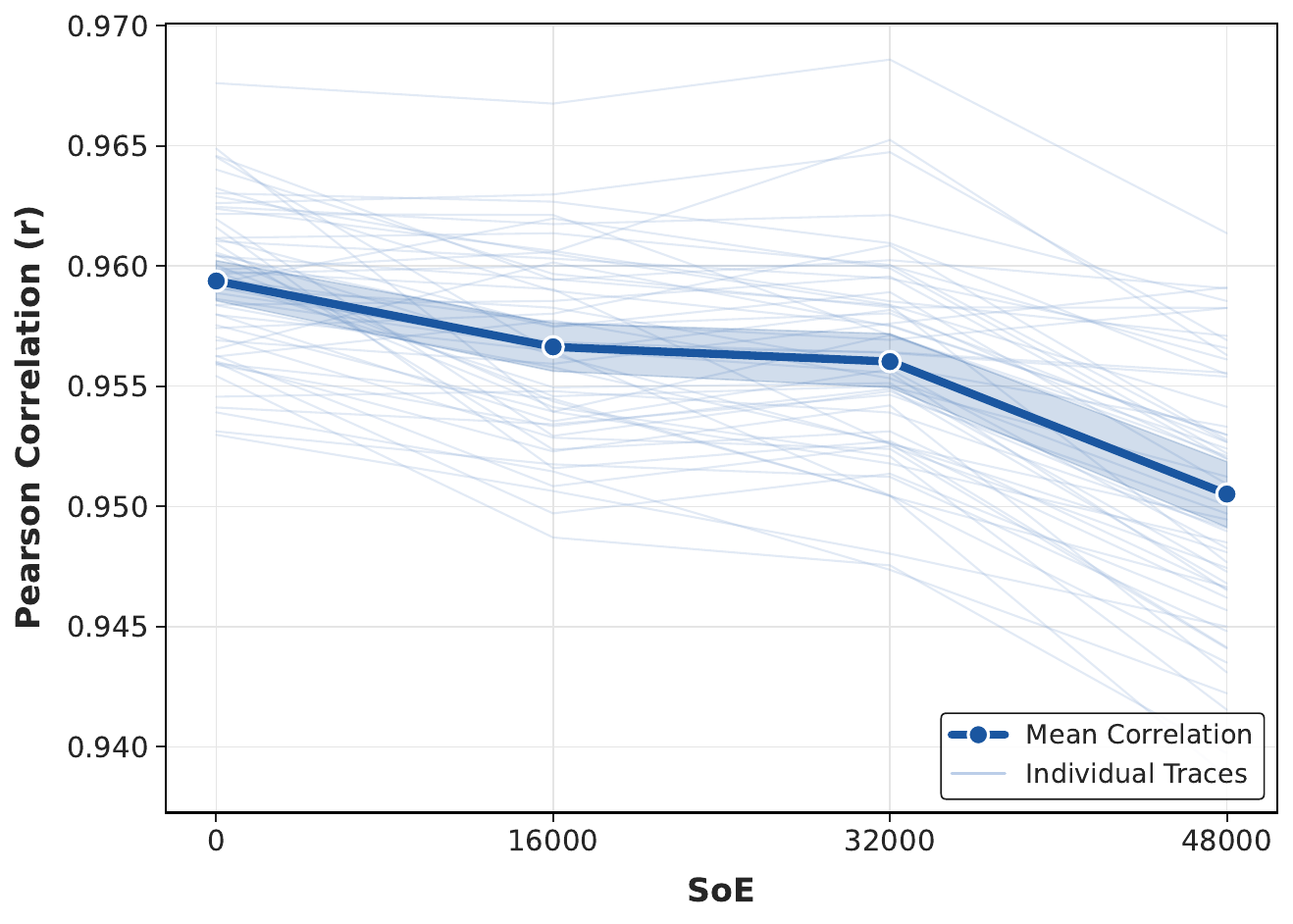}
    \caption{Pairwise Pearson correlation of English surprisal values between all L1 bilingual models across the SoE stages. The dark blue line represents the mean correlation, the shaded band shows 95\% confidence interval, and the light blue lines show individual pairwise comparisons.}
    \label{fig:surprisal_corr}
\end{figure}

Because all bilingual models in this study are trained and evaluated on English, their surprisal estimates for English text share substantial variance. We calculate the pairwise Pearson correlation of per-word surprisal across all L1 bilingual models using the 12 English texts from the MECO dataset. Figure \ref{fig:surprisal_corr} confirms this high overlap, showing that all pairwise correlations remain above 0.94. However, the mean correlation decreases slightly as SoE increases (from 0.96 to 0.95), suggesting that extended L1 training introduces small but noticeable L1-specific differences between models.

\section{Proficiency Analysis}

\subsection{Isolated effects of Bilingual and Monolingual Surprisal}
\label{sec:absolute_surprisal}

\begin{figure*}[h]
    \centering
    \includegraphics[width=\textwidth]{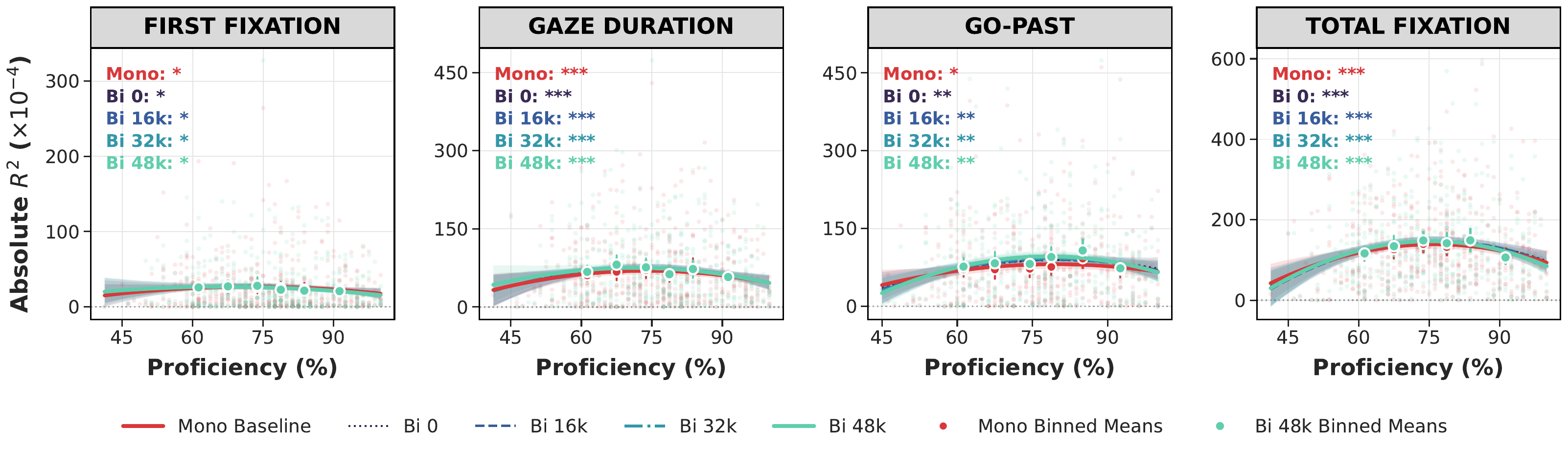}
    \caption{Explanatory power ($\Delta R^2$ over the baseline, scaled by $10^4$) of monolingual and of bilingual surprisal across readers' English proficiency (LexTALE \%) in four reading measures. Faint dots indicate individual observations, and the horizontal dotted line denotes zero gain. Inset labels indicate the significance of the quadratic proficiency term in a WLS with L1-subset fixed effects and standard errors clustered by L1 subset ($^*p < 0.05$, $^{**}p < 0.01$, $^{***}p < 0.001$, $^{\text{ns}}\text{not significant}$). Both bilingual and monolingual models exhibit a pronounced inverted-U relationship with proficiency.}
    \label{fig:absolute_r2}
\end{figure*}

In \S\ref{sec:proficiency_analysis}, we evaluated the unique advantage of bilingual surprisal over monolingual surprisal ($\Delta R^2 = R^2_{\text{bilingual}} - R^2_{\text{monolingual}}$). To provide full context for this metric, Figure \ref{fig:absolute_r2} displays the  explanatory power of  the monolingual and the bilingual surprisal models separately over the baseline (which contains only surface covariates such as word length and frequency). 

The absolute surprisal from both monolingual and bilingual models follows a highly similar inverted-U trajectory across proficiency levels, peaking for intermediate readers. This replicates the findings of \citet{berzak2023}, confirming that intermediate L2 readers rely most heavily on context-dependent expectations. Because the bilingual and monolingual LMs share a vast majority of their English linguistic representations (as evidenced by their high surprisal correlation), their individual lines overlap almost entirely. Crucially, in the late processing measures (go-past time and total fixation duration), the bilingual models consistently diverge from and peak slightly above the monolingual baseline. The subtraction methodology utilized in \S\ref{sec:proficiency_analysis} isolates this specific gap, confirming that the isolated cross-linguistic transfer effect also scales non-linearly with reader proficiency.

\subsection{Quadratic Relationship Evaluation}
\label{sec:quadratic_estimation}
Table \ref{tab:proficiency_model_comparison} provides the statistical evaluation comparing linear and quadratic specifications for the effect of proficiency on the bilingual advantage ($\Delta R^2$). For early measures (first fixation duration and gaze duration), the quadratic term is not statistically significant, and the negative $\Delta$AIC values indicate that the simpler linear baseline is generally preferred. In contrast, for late measures (go-past time and total fixation duration), the quadratic model provides a better fit ($p_{\text{quad}} < 0.05$ and $\Delta\text{AIC} > 2.0$) for models with SoE $\ge$ 16k. The negative $\beta_{\text{quad}}$ coefficients align with the observed inverted-U trajectory. Support for this non-linear relationship tends to increase with longer L1 training; in go-past time, for instance, the $\Delta$AIC rises from +0.6 at SoE 0 to +9.5 at SoE 48k. This suggests that a stronger L1 influence on the L2 (due to extended L1 training and delayed L2 exposure) increases the predictive value of bilingual surprisal mostly for intermediate learners.

\begin{table}[t]
\centering
\small
\setlength{\tabcolsep}{4.5pt}
\begin{tabular}{llcccc}
\toprule
\textbf{Metric} & \textbf{SoE} & $N$ & $\beta_{\text{quad}}$ & $p_{\text{quad}}$ & $\Delta\text{AIC}$ \\
\midrule
First Fixation   & 0          & 480 & +0.002 & 0.464 & $-$1.5 \\
                 & 16k                    & 480 & +0.002 & 0.494 & $-$1.7 \\
                 & 32k                    & 480 & $-$0.001 & 0.793 & $-$2.0 \\
                 & 48k            & 480 & +0.003 & 0.436 & $-$1.5 \\
\midrule
Gaze Duration    & 0          & 480 & $-$0.007 & 0.129 & $-$0.2 \\
                 & 16k                    & 480 & $-$0.007 & 0.093 & +0.0 \\
                 & 32k                    & 480 & $-$0.009 & 0.053 & +0.8 \\
                 & 48k            & 480 & $-$0.002 & 0.703 & $-$1.9 \\
\midrule
Go-Past     & 0          & 434 & $-$0.010 & 0.055 & +0.6 \\
                 & \textbf{16k}           & 434 & $-$0.014 & 0.011 & +2.5 \\
                 & \textbf{32k}           & 434 & $-$0.022 & $< 0.001$ & +6.9 \\
                 & \textbf{48k}   & 434 & $-$0.029 & $< 0.001$ & +9.5 \\
\midrule
Total Fixation   & 0          & 480 & $-$0.011 & 0.074 & +0.9 \\
                 & \textbf{16k}           & 480 & $-$0.017 & 0.002 & +4.5 \\
                 & \textbf{32k}           & 480 & $-$0.018 & 0.003 & +4.6 \\
                 & \textbf{48k}   & 480 & $-$0.022 & 0.004 & +5.0 \\
\bottomrule
\end{tabular}
\caption{Model comparison evaluating non-linear proficiency moderation across eye-tracking measures. $\beta_{\text{quad}}$ denotes the estimated coefficient of the quadratic proficiency term, and $p_{\text{quad}}$ denotes its robust HC3 significance. $\Delta\text{AIC} = \text{AIC}_{\text{linear}} - \text{AIC}_{\text{quadratic}}$, where values $> +2.0$ indicate that the quadratic model improves on the linear baseline by more than the penalty for its extra parameter \citep{burnham2002model}. Boldface marks exposure stages where both criteria favor the quadratic model ($p_{\text{quad}} < 0.05$, $\Delta\text{AIC} > 2.0$).}
\label{tab:proficiency_model_comparison}
\end{table}

\subsection{Robustness to Composite Proficiency (PCA)}
\label{sec:proficiency_pca}
\begin{figure*}[t]
  \centering
  \includegraphics[width=\textwidth]{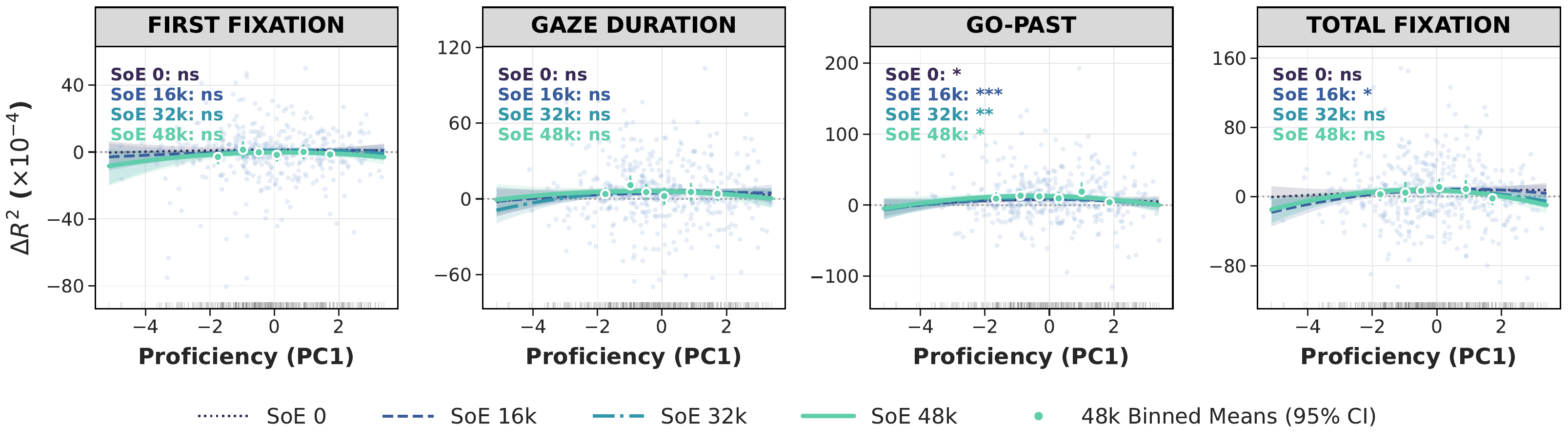}
  \caption{Bilingual surprisal advantage ($\Delta R^2 = R^2_{\text{bilingual}} - R^2_{\text{monolingual}}$, scaled by $10^4$) modeled against composite L2 proficiency (PC1 extracted across LexTALE, Vocabulary, Spelling, TOWRE word, and TOWRE nonword; with 52.5\% variance explained) in four reading measures. Curves depict second-order polynomial regressions across SoEs; markers denote empirical binned means ($\pm 95\%$ CI) at SoE=48k. Inset labels indicate the significance of the quadratic proficiency term in a WLS with L1-subset fixed effects and standard errors clustered by L1 subset ($^*p < 0.05$, $^{**}p < 0.01$, $^{***}p < 0.001$, $^{\text{ns}}\text{not significant}$). Faint points represent individual participants, and the dotted horizontal line marks zero gain.}
  \label{fig:proficiency_pca}
\end{figure*}

To verify that the non-linear proficiency moderation observed with LexTALE is not an artifact of this single metric, we evaluated whether these findings generalize across other proficiency measures reported in MECO \cite{Kuperman2025}. Specifically, we extracted a composite index via Principal Component Analysis (PCA) across five standardized assessments (LexTALE, vocabulary knowledge, spelling, TOWRE-2 test of words and nonwords), with the first principal component (PC1) accounting for $52.5\%$ of the total variance.

As shown in Figure~\ref{fig:proficiency_pca}, modeling $\Delta R^2$ against PC1 closely mirrors the findings obtained with LexTALE alone. In early reading measures (first fixation duration and gaze duration), the quadratic term remains non-significant across all SoE (p>0.05), confirming our earlier findings using LexTALE. In contrast, late processing in go-past time robustly replicates the inverted-U trajectory across all SoE (p<0.05). For total fixation duration, the quadratic trajectory exhibits the same qualitative inverted-U curvature, although under strict L1-subset controls its statistical significance is attenuated relative to LexTALE, reaching significance only at SoE=16k. Overall, these results demonstrate that the bilingual processing advantage peaking among intermediate-proficiency readers in late-stage eye-movement measures generalizes across alternative proficiency assessments, particularly for go-past time.

\end{document}